%% file: representing-crf.tex
\documentclass[pdflatex, sn-mathphys, iicol]{sn-jnl}

\jyear{2022}%

\theoremstyle{thmstyleone}%
\theoremstyle{thmstyletwo}%

\theoremstyle{thmstylethree}%

\begin{document}

\graphicspath{{figures/}}

\title[Representing Camera Response Function]{Representing Camera Response Function by a Single Latent Variable and Fully Connected Neural Network}


\author[1]{\fnm{Yunfeng} \sur{Zhao}}\email{yzhao25@qub.ac.uk}

\author[1]{\fnm{Stuart} \sur{Ferguson}}\email{r.ferguson@ee.qub.ac.uk}

\author[2]{\fnm{Huiyu} \sur{Zhou}}\email{hz143@leicester.ac.uk}

\author*[1]{\fnm{Karen} \sur{Rafferty}}\email{k.rafferty@qub.ac.uk}

\affil*[1]{\orgdiv{School of Electronics, Electrical Engineering \& Computer Science}, \orgname{Queen's University Belfast}, \orgaddress{\street{16A Malone Road}, \city{Belfast}, \postcode{BT9 5BN}, \country{United Kingdom}}}

\affil[2]{\orgdiv{School of Computing and Mathematical Sciences}, \orgname{University of Leicester}, \orgaddress{\street{University Road}, \city{Leicester}, \postcode{LE1 7RH}, \country{United Kingdom}}}


\abstract{Modelling the mapping from scene irradiance to image intensity is essential for many computer vision tasks. Such mapping is known as the camera response. Most digital cameras use a nonlinear function to map irradiance, as measured by the sensor to an image intensity used to record the photograph. Modelling of the response is necessary for the nonlinear calibration. In this paper, a new high-performance camera response model that uses a single latent variable and fully connected neural network is proposed. The model is produced using unsupervised learning with an autoencoder on real-world (example) camera responses. Neural architecture searching is then used to find the optimal neural network architecture. A latent distribution learning approach was introduced to constrain the latent distribution. The proposed model achieved state-of-the-art CRF representation accuracy in a number of benchmark tests, but is almost twice as fast as the best current models when performing the maximum likelihood estimation during camera response calibration due to the simple yet efficient model representation.}




\maketitle

\section{Introduction}\label{sec1}

A camera response function (CRF) describes the mapping between the radiant energy received by an image sensor and the intensity output of a camera in the final images \cite{Ng2009}. Most cameras are manufactured with nonlinear CRFs \cite{Grossberg2004}. Such nonlinearity is introduced during the stages of image formation in the camera. For instances, analog-digital conversion in the image sensor, white balance adjustment that minimises image colour drift due to differing illumination, gamma correction that expands the luminance range to be interpreted, and tone mapping for optimising the image visual quality \cite{Fu2011}. Popular CRF models include Empirical Model of Response (EMoR), generalized gamma curve model (GGCM), polynomial, and gamma \cite{Grossberg2004, Ng2007}.

Calibration of a camera response is crucial in many computer vision tasks. Examples of such tasks include image mosaicing where multiple images need to be flawlessly coupled together \cite{Kim2008}, high dynamic range imaging where images of multiple exposures are used to produce images with greater dynamic ranges \cite{Mann2000}, and denoising that removes motion blur \cite{Tai2013}. CRF calibration also has application in digital forensics \cite{Ng2009}. 

Elaborate CRF representation modelling is the foundation for accurate and rapid CRF calibration. The calibration can be seen as an optimisation process where often the optimal parameters of a selected CRF representation model are calculated to best describe the camera response. The existing CRF models are mostly parametric with multiple parameters. The solution spaces for optimising these parameters are complex with arbitrary distributions. Thus, it takes a long time to calibrate the optimal parameters using existing models.

In this paper, a novel and high-performance non-deterministic CRF representation model, the Single Latent Representation model (SLR), is proposed based on the autoencoder, neural architecture search (NAS), and latent distribution learning (LDL) techniques. This work presents the following contributions. 1) Pattern of real-world CRFs were extracted by unsupervised learning and represented by a single latent variable using autoencoder. 2) Two approaches (i.e., a LDL and a supervised learning approach using handcrafted feature) are proposed and applied during model representation learning to constrain the latent distribution which further improves the accuracy of camera calibration. 3) A naïve NAS algorithm is used to seek for the optimal autoencoder architecture considering both model accuracy and complexity. 4) The proposed model achieves state-of-the-art performance in terms of accuracy and efficiency of CRF modelling but executes in less than half the time than current best algorithms during CRF calibration.

\section{Related work}\label{sec2}

The latest successful CRF representation model is perhaps the EMoR by Grossberg and Nayar proposed in 2004 \cite{Grossberg2004}. The EMoR describes a CRF by linearly composing a collection of principal components or eigenvectors generated by applying Principal Component Analysis (PCA) on 201 real-world CRFs known as the Database of Response Function (DoRF). Each CRF curve is composed by $1024$ uniformly sampled irradiance-intensity converting ratios and is normalised between and passes through $\left(0,0\right)$ and $\left(1,1\right)$. By EMoR, an approximation $\widetilde{f}$ to the CRF $f$ can be constructed from $k$ coefficients and the corresponding eigenvectors:

\begin{equation}\label{equ:emor}
   \widetilde{f}=f_0+\bm{c}_k^T \bm{H}_k
\end{equation}
where $f_0$ is the base function calculated by averaging all the CRFs in DoRF, $\bm{c}_k=\bm{H}_k^T\left ( f-f_0\right )$ is the model coefficients, and $\bm{H}_k:=\left [ \bm{h}_1\cdots \bm{h}_k\right ]$ is the first $k$ eigenvectors with the largest eigenvalues.

EMoR is an efficient model to represent CRFs by a very small number of parameters or coefficients. As reported in the paper \cite{Grossberg2004}, three eigenvalues encode 99.5 percent of the cumulative energies associated with the eigenvalues in DoRF. So far, it is the most widely adopted model for representing a CRF due to its high representing accuracy and simplicity \cite{Fu2011, Li2017, Lin2005, Bergmann2018, Matsushita2007, Seon2008, Kim2008}.

Polynomial and gamma curves are the two other popular models used for CRF representation whose performances are slightly worse than the EMoR according to a benchmark \cite{Ng2007}. A high-order polynomial has the general form:

\begin{equation}\label{equ:polynomial}
   f_{\bm{\omega}} \left ( \bm{x} \right )=\sum_{i=1}^{M}{\bm{\omega}}_i \bm{x}^i
\end{equation}
where $M$ and $\bm{\omega}$ are the order number and model coefficients, respectively, and they are the parameters to be determined through camera calibration. $\bm{x}\in \left [ 0,1\right ]$ is the model input and represents image pixel intensity. 

In general, gamma curves follow the basic form:

\begin{equation}\label{equ:gamma}
   f\left ( \bm{x} \right )=\bm{x}^\gamma 
\end{equation}
where $\gamma$ is the gamma value typically determined through calibration. This model has been applied in numeral works \cite{Zhao2018, Farid2001}.

An extended version of gamma curve named GGCM has been proposed \cite{Ng2007} and applied \cite{Rodrigues2015}. It is denoted in (\ref{equ:ggcm}).

\begin{equation}\label{equ:ggcm}
   f_{\bm{\omega}} \left ( \bm{x} \right )=\bm{x}^{P_{\bm{\omega}} \left ( \bm{x}\right )}
\end{equation}
where the gamma value in the basic form is replaced by a polynomial term $P_{\bm{\omega}} \left ( \bm{x}\right )=\sum_{i=0}^{N}{\bm{\omega}}_i\bm{x}^i$.

A limitation of the current CRF representation models is certainly the high dimensional and complex solution space for solving the optimal model parameters during calibration. A CRF representation with a minimum number of model parameters, e.g., the single gamma value for gamma curves, is demanded for simplifying the calibration. Autoencoder has the ability of generalising and has been used for representation modelling \cite{CHARTE202093}. It compresses data into a much lower dimensional latent space represented by a few latent variables and provides a potential solution to CRF representation. However, such work has not been reported yet.

In general, autoencoder is a neural network that consists of an encoder and a decoder. The encoder maps the input data $\bm{x}$ to a latent representation $\bm{z}$. And the decoder reconstructs $\bm{z}$ back to the input data $\tilde{\bm{x}}$. The latent representation and the model weights are trained by minimising the difference between the input and reconstructed data in an unsupervised process \cite{hinton1994}. 

In the work by Makhzani et al. \cite{Makhzani2015}, Adversarial Autoencoder (AAE) has been introduced combining autoencoder with generative adversarial training to deliver unsupervised learning on multiple objectives. It can impose a constraint on the latent distribution by the adversarial training process. The value function of adversarial training can be represented as:

\begin{equation}\label{equ:adversarial-training}
    \begin{aligned}
        \min_{G} \max_{D} V\left ( D, G \right ) & = \mathbb{E}_{\bm{x} \sim p_d}\left [ \log D\left (\bm{x}\right )\right ] \\
        & + \mathbb{E}_{\bm{z} \sim p\left(\bm{z}\right)}\left [ \log \left ( 1 - D \left ( G \left ( \bm{z} \right )\right ) \right ) \right ]
    \end{aligned}
\end{equation}
where the encoder in the autoencoder also acts as the generator $G$ in adversarial network to produce the latent representations $\bm{z}$ from the data distribution $P_d$. And at the same time, a discriminator $D$ calculates the probability that a representation is generated from the data or prior distribution. AAE has been successfully applied in applications such as image anomaly detection \cite{Beggel2020} and classification \cite{Makhzani2015}.

A Variational Autoencoder (VAE) \cite{kingma2014-VAE} is another popular autoencoder model capable of constraining the distribution of latent representation. Such constraint is achieved by a recognition network that predicts the posterior distribution of the latent space.

Recent advancement of neural networks for end-to-end feature representation and data processing increases the demand for automating the architecture engineering which is time-consuming and usually done manually. NAS automates the neural network engineering process. It can be summarized as three topics: search space, search strategy, and performance evaluation strategy \cite{Elsken2018}. Search space defines the architecture searching scope and usually involves human prior knowledge. Search strategy determines how the search space is explored. And performance evaluation strategy quantifies candidate model performance.

\section{Proposed method}\label{sec5}

\subsection{Autoencoder-based CRF representation model}

\begin{figure}[h]%
    \centering
    \includegraphics[width=1.0\linewidth]{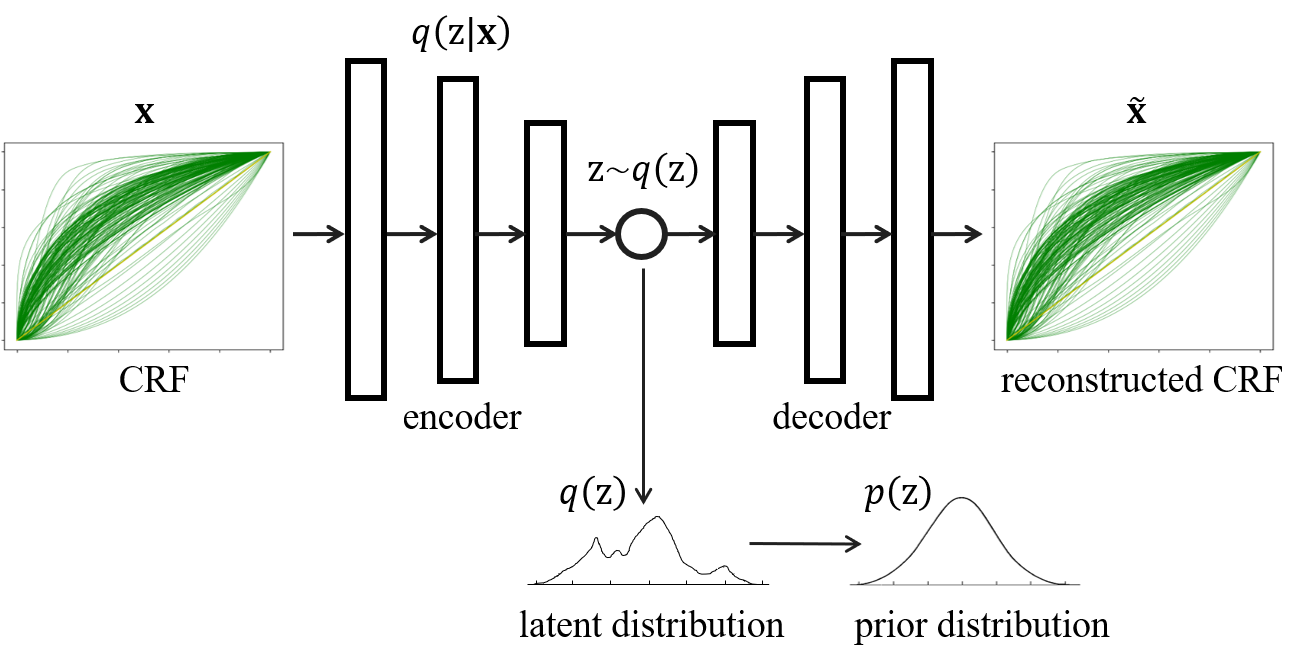}
    \caption{Architecture of the proposed model for CRF representation. The top row represents a multi-layer fully connected autoencoder with a single latent variable. The bottom row demonstrates the latent distribution and an objective prior distribution.}\label{fig1}
\end{figure}

As shown in Fig. \ref{fig1}, the proposed Single Latent variable camera response Representation (SLR) model inputs a CRF represented by 1,024 uniformly sampled points on the function, reduces the dimensionality to the latent space by the encoder, and outputs the reconstructed CRF by the decoder. As a result, a CRF can be represented by the latent variables   in the latent space of the proposed model. A multi-layer fully connected autoencoder with the same number of input and output neurons is selected as the representation model of CRFs. 

In our model, the number of hidden layers in either the encoder or decoder is denoted by $L$. Both the encoder and decoder contain either one, two, or three hidden layers, i.e., $L \in \left\{ {1,2,3} \right\}$. Each hidden layer contains varied number of neurons, denoted by ${C_l}$ where $l \in \left\{ {1, \ldots ,L} \right\}$ is the layer index. ${C_z}$ is the number of latent variables in the model. Dropout operation is added to prevent the model from overfitting \cite{Nitish2014}. Nonlinearity is introduced by an activation function on each unit. The feed-forward operation of the proposed model has the form: 

\begin{equation}
    \begin{split}
        {r_j}&\sim{\rm{Bernoulli}}\left( p \right) \\
        {{\tilde u}^{\left( l \right)}} &= {r^{\left( l \right)}} * {u^{\left( l \right)}} \\
        v_i^{\left( {l + 1} \right)} &= w_i^{\left( {l + 1} \right)}{{\tilde u}^l} + b_i^{\left( {l + 1} \right)} \\
        u_i^{\left( {l + 1} \right)} &= g\left( {v_i^{\left( {l + 1} \right)}} \right)
    \end{split}
\end{equation}
where ${r^{\left( l \right)}}$ is a vector of independent Bernoulli random variables with each element a probability $p$ of being 1, * denotes an element-wise product, ${u^{\left( l \right)}}$ denotes the output vector calculated from the input vector ${v^{\left( l \right)}}$ into layer $l$, ${w^{\left( l \right)}}$ and ${b^{\left( l \right)}}$ are the model weights and bias at layer $l$, $g$ is the activation function. 

The output vector from layer $l$ is firstly sampled by the dropout operation and then processed by the weights and bias. The processed outputs are nonlinearly activated and used as inputs to the next layer. This process is repeated layer by layer. At test time, the model weights are scaled by $p$ to infer without the effect of the dropout. For CRF construction, the latent variable is used as the input, and the reconstructed CRF $\tilde x$ can be obtained at the final output layer.

The model weights in the autoencoder are learnt by independently back-propagating the gradients calculated from the derivatives of the losses. The reconstruction loss is the mean-square-error (MSE) between the input $x$ and reconstructed CRFs $\tilde x$:

\begin{equation}\label{equ:mse}
    MSE\left( {x,\tilde x} \right) = \frac{1}{N}\sum\limits_{i = 1}^N {{{\left( {{x_i} - {{\tilde x}_i}} \right)}^2}}
\end{equation}
where $N$ is the number of training data. Meanwhile, a smoothness loss is imposed on the reconstructed CRF $\tilde x$ as a CRF is usually a smooth and continuous function based on the observation from the CRFs in the DoRF:

\begin{equation}\label{equ:loss}
    {\cal L}\left( {\tilde x} \right) = {\left\| {{{\tilde x}^\prime }} \right\|_2}
\end{equation}
where $\tilde x$ is the first derivative of the reconstructed CRF, ${\parallel _2}$ denotes the l2-norm.

The optimal number of hidden layers and number of neurons in each hidden layer are determined by NAS.

\subsection{Naïve neural architecture search}

The optimal architecture of the proposed SLR model in terms of both model accuracy and complexity is determined by NAS. NAS not only helps find the desired model architecture but also brings flexibility to the model design (e.g., when an extension of the number of latent variables is needed). Since devices with relatively limited computing resources such as mobile phones are being considered for running the proposed model, the model performance estimation needs to be taken cognisance of both the model complexity and its accuracy.

The search space defined is three hidden layers with optional neuron numbers ${h_1} = \left[ {10,20,50,100,200,500} \right]$, ${h_2} = \left[ {0,10,20,50,100,200} \right]$, and ${h_3} = \left[ {0,10,20,50,100} \right]$ for both encoder and decoder. Note that when hidden layer two has no neuron, hidden layer three must also have no neuron.

We aim to minimise the model complexity while maximising its accuracy. However, balancing the trade-offs between model complexity and accuracy is a persistent challenge in NAS.

In this paper, the optimal model architecture is determined by a newly proposed NAS method named naïve NAS. Initially, the naïve NAS searches for possible neural architecture in the search space. It then selects $M$ candidate architectures with the highest accuracies from the search. Eventually, the optimal architecture is chosen from those $M$ candidates with the lowest model complexity. The naïve NAS is illustrated in Algorithm \ref{alg:naive-nas}.

\begin{algorithm}[htbp]
    \caption{Na\"ive neural architecture search algorithm}
    \label{alg:naive-nas}
    \begin{algorithmic}[1]
        \Require
        $\mathcal{A}$ is the architecture search space. $M$ is the number of architectures with the highest accuracies to be selected from $\mathcal{A}$ for the complexity estimation.
        \Ensure
        The optimal architecture $A \in \mathcal{A}$
        \State Construct a Set vector $S$ with length $N$ which equals to the number of elements in $\mathcal{A}$
        \For { $k \leftarrow 1$ to $N$ }
        \State Obtain the architecture object from the search space $A \leftarrow \mathcal{A}_k$
        \State Estimate the architecture accuracy $a \leftarrow Accuracy\left(A\right)$
        \State Estimate the architecture complexity $c \leftarrow Complexity\left(A\right)$
        \State Construct and store a Set object $S_k \leftarrow \left\{a, c, A\right\}$
        \EndFor
        \State Descending order $S$ based on the accuracy metric $a$
        \State Select $M$ candidate architectures with the highest accuracies $S' \leftarrow S\left[0:M\right]$
        \State Ascending order $S'$ based on the complexity metric $c$
        \State Obtain the architecture object $A$ from the first element of the ordered $S'$
        \\
        \Return $A$
    \end{algorithmic}
\end{algorithm}

Existing variable search strategies can be coupled with the proposed naïve NAS. The Grid Search \cite{Liashchynskyi2019} is the selected strategy since it is thorough and the proposed model is light-weight (i.e., the performance estimation of each candidate architecture can be completed in less than a minute) and the search space is discrete and small (i.e., with only a total of 156 valid candidate architectures in the search space). 

The model complexity is calculated by the total number of weights in either the encoder or decoder of the SLR with considering the latent variable:

\begin{equation}\label{equ:complexity}
    Complexity\sim\left[ {\left( {\sum\limits_{l = 1}^L {{C_{l - 1}}{C_l} + {C_l}} } \right) + {C_L}{C_z} + {C_z}} \right]
\end{equation}
where $L$ is the total number of layers in the encoder or decoder, $C$ is the number of neurons in a specific layer, and ${C_z}$ is the number of latent variables in the model.
The model accuracy is measured by a three-fold cross-validation and (\ref{equ:mse}):

\begin{equation}\label{equ:complexity}
    Accuracy \sim MSE\left( {x,\tilde x} \right)
\end{equation}
where $x$ and $\tilde x$ are the reconstructed and validated CRF curves.

\subsection{Constraint on the latent distribution}

The latent variable $z$ in the autoencoder follows an arbitrary distribution by default. Two approaches (i.e., a distribution learning and a supervised learning approach using heuristics) have been proposed to constrain the latent variable to follow a prior distribution to help the optimisation process to more accurately find the best $z$ during calibration. 

In the first approach, the latent distribution is constrained by "learning" from the objective distribution. It is named \emph{latent distribution learning} (LDL) and achieved by minimising the Kullback-Leibler Divergence (KL-divergence) between the latent and objective distributions. 

The latent distribution is approximated by a normal distribution $y\sim{\cal N}\left( {\mu ,\sigma } \right)$. The normal distribution maximum likelihood of the latent variable is estimated by:

\begin{equation}
    \begin{split}
        \mu  &= \bar y,\\
        {\sigma ^2} &= \sum\limits_{i = 1}^N {{{\left( {{y_i} - \mu } \right)}^2}} 
    \end{split}
\end{equation}
where $y$ is the samplings on the latent distribution and $M$ is the number of samplings.

KL-divergence between two normal distributions has the form:

\begin{equation}
    \begin{split}
        KL&\left( {{{\cal N}_1}\left( {{\mu _1},{\sigma _1}} \right),{{\cal N}_2}\left( {{\mu _2},{\sigma _2}} \right)} \right)\\
        &=  - \int {{{\cal N}_1}\log \left( {{{\cal N}_2}} \right)} dy + \int {{{\cal N}_1}\log \left( {{{\cal N}_1}} \right)} dy\\
        &= \log \left( {\frac{{{\sigma _2}}}{{{\sigma _1}}}} \right) + \frac{{\sigma _1^2 + {{\left( {{\mu _1} - {\mu _2}} \right)}^2}}}{{2\sigma _2^2}} - \frac{1}{2}
    \end{split}
\end{equation}

The KL-divergence between the estimated latent distribution ${\cal N}\left( {{\mu _1},{\sigma _1}} \right)$ and the objective standard normal distribution ${\cal N}\left( {0,1} \right)$ can be simplified to:

\begin{equation}
    \begin{split}
        KL&\left( {{\cal N}\left( {{\mu _1},{\sigma _1}} \right),{\cal N}\left( {0,1} \right)} \right) \\
        &= \frac{1}{2}\left( {\mu _1^2 + \sigma _1^2 - 2\log {\sigma _1} - 1} \right)
    \end{split}
\end{equation}

This is used as the cost for the latent distribution learning in the proposed SLR model.
The second approach, named AUC, generates a label for each CRF as the true latent value for distribution constraining. The label is generated by a so-called "area under curve" approach which calculates the area between the CRF and diagonal curve:

\begin{equation}
    \iota  = \sum\limits_{i = 1}^N {\left( {{x_i} - \frac{i}{N}} \right)}
\end{equation}
where $N$ is the number of samplings in each CRF curve, which is 1,024 for those in the DoRF. The latent distribution is trained using supervised learning by minimising the MSE between the latent and true values.

\section{Experiments and results}\label{exp}

This section details the experimental setup used to examine and test the proposed model. All the processing and evaluations were performed on a laptop computer with a 2.6-GHz Intel Core i7 processor and a 16-GB memory. To accelerate the optimisation process, a NVIDIA GeForce RTX 2060 GPU was employed.

\subsection{Datasets}

\begin{figure}[htbp]
    \centering
    \includegraphics[width=0.45\textwidth]{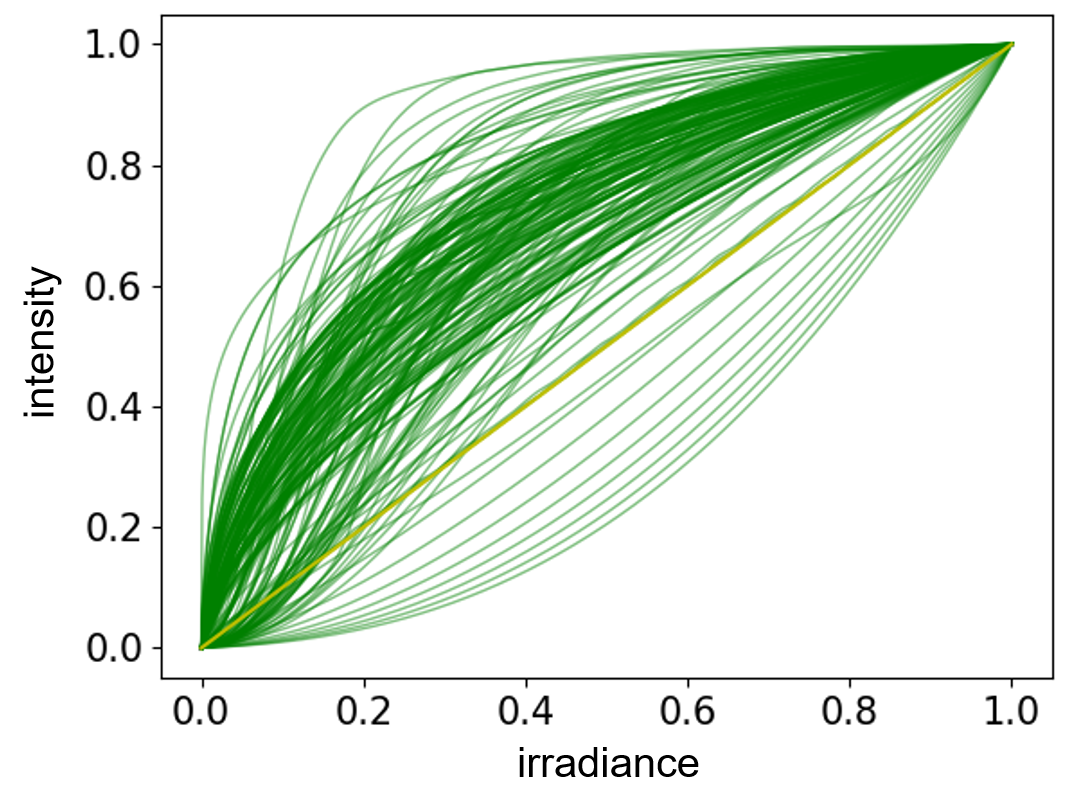}
    \includegraphics[width=0.45\textwidth]{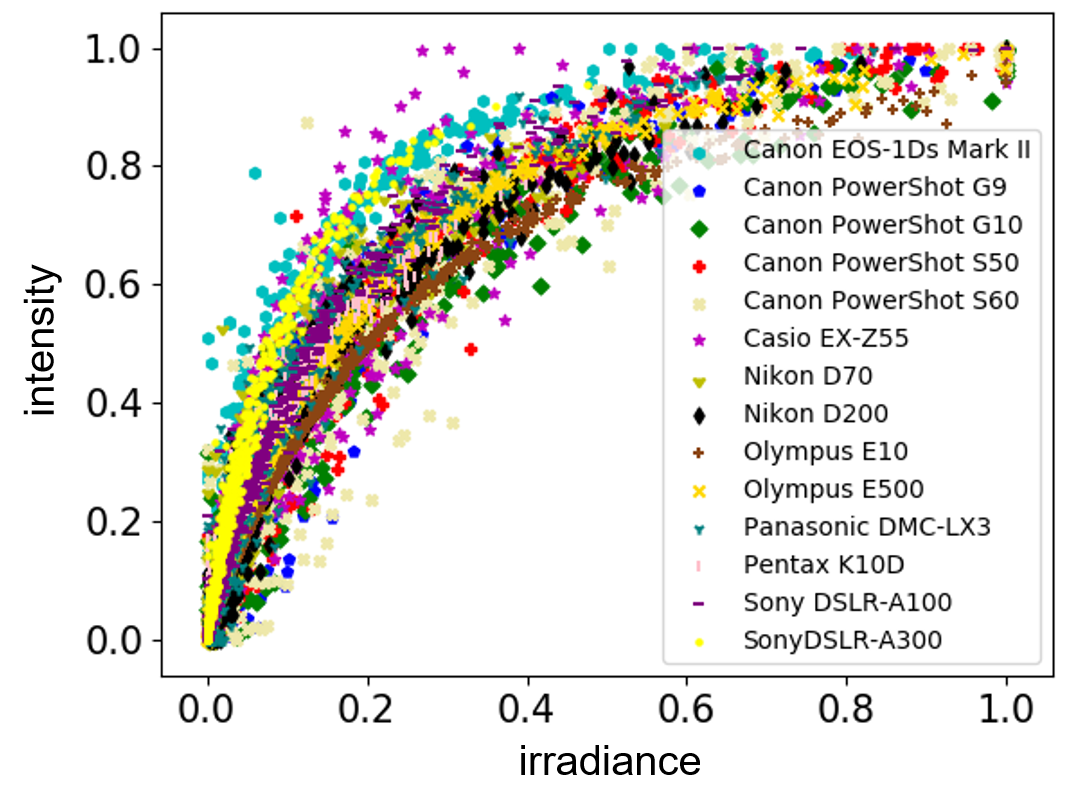}
    \caption{Distribution of the two datasets used for model validation. (a) The CRFs of 201 real-world cameras in the DoRF. Each green curve represents a CRF in the database. (b) Irradience-intensity scatter plot of the CCPs extracted from the 14 cameras selected from the Middlebury dataset. CCPs of different cameras were rendered in varied colours.}
    \label{fig:datasets}
 \end{figure}

Two datasets, i.e., the DoRF and a modified Middlebury \cite{chakrabarti2009}, were prepared for performing the validations and benchmarks. Data distribution of these two datasets are demonstrated in Fig. \ref{fig:datasets}.

The DoRF contains 201 CRFs and is currently the most comprehensive dataset of CRFs produced from real-world camera models. This dataset was used in our experiments without modification.

The modified Middlebury dataset contains a total of 112 images. Images of 14 cameras were selected from the original dataset. These cameras were chosen because of their higher cross-channel response uniformity. Each of these cameras took eight images of a Macbeth colour chart under two uniform illuminations and four fixed exposures. This dataset provides an abundance of variation for evaluating CRF calibration accuracy.

The colour patch (CP) locations in the images in the second dataset (24 CP for each image) were carefully labelled by utilising a custom-developed Python script so that the CPs can be extracted and aligned with each other across different images. The true colour values of the CPs are extracted from the RAW images.

\subsection{Evaluation metrics}

The root-mean-square error (RMSE) \cite{Grossberg2004, Li2017, Lin2004, Sharma2020, Gehler2008} has been widely used to quantify colour difference. It measures the Euclidean distance between two compared vectors:

\begin{equation}
    d\left( {{u_i},{v_i}} \right) = \sqrt {\frac{1}{N}\sum\limits_{i = 1}^N {{{\left( {{u_i} - {v_i}} \right)}^2}} }
\end{equation}
where $u$ and $v$ are the compared vectors and $N$ is the number of items in each of the vectors. A smaller RMSE indicates a better result. A 0 RMSE illustrates identical results.

In the experiments, the RMSEs calculated from comparing the reconstructed CRF with CRFs in the DoRF in the first experiment or those from comparing colour values of JPG and corresponding RAW images in the second experiment were collected into a result vector $h$ for statistical analysis:

\begin{equation}
    h = \left. {\left[ {\begin{array}{*{20}{c}}
        {{e_0}}\\
         \vdots \\
        {{e_{ - 1}}}
        \end{array}} \right]} \right\}C
\end{equation}
where $C$ is the number of camera models to be compared.

The Mean of the result vector $h$ was used as the overall performance evaluation indicator in the first experiment. In the second experiment, five metrics are used to evaluate the result vector produced by each method. The first four are statistical metrics (i.e., Mean, Standard Derivation, Maximum, and 95 Percentile) that reflect model accuracy. Among these four metrics, the Mean of $h$ can be seen as the overall performance metric for accuracy. The time metric was evaluated as the total time needed in seconds (s) for calibrating all the 14 camera models in the second dataset. We considered $\Delta RMSE > 0.005$ to be the thresholds for significant performance difference in the second experiment.

\subsection{Latent distribution constraint benchmarks}

\begin{figure}[htbp]
    \centering
    \subfloat[The objective latent distribution]{\includegraphics[width=0.45\linewidth]{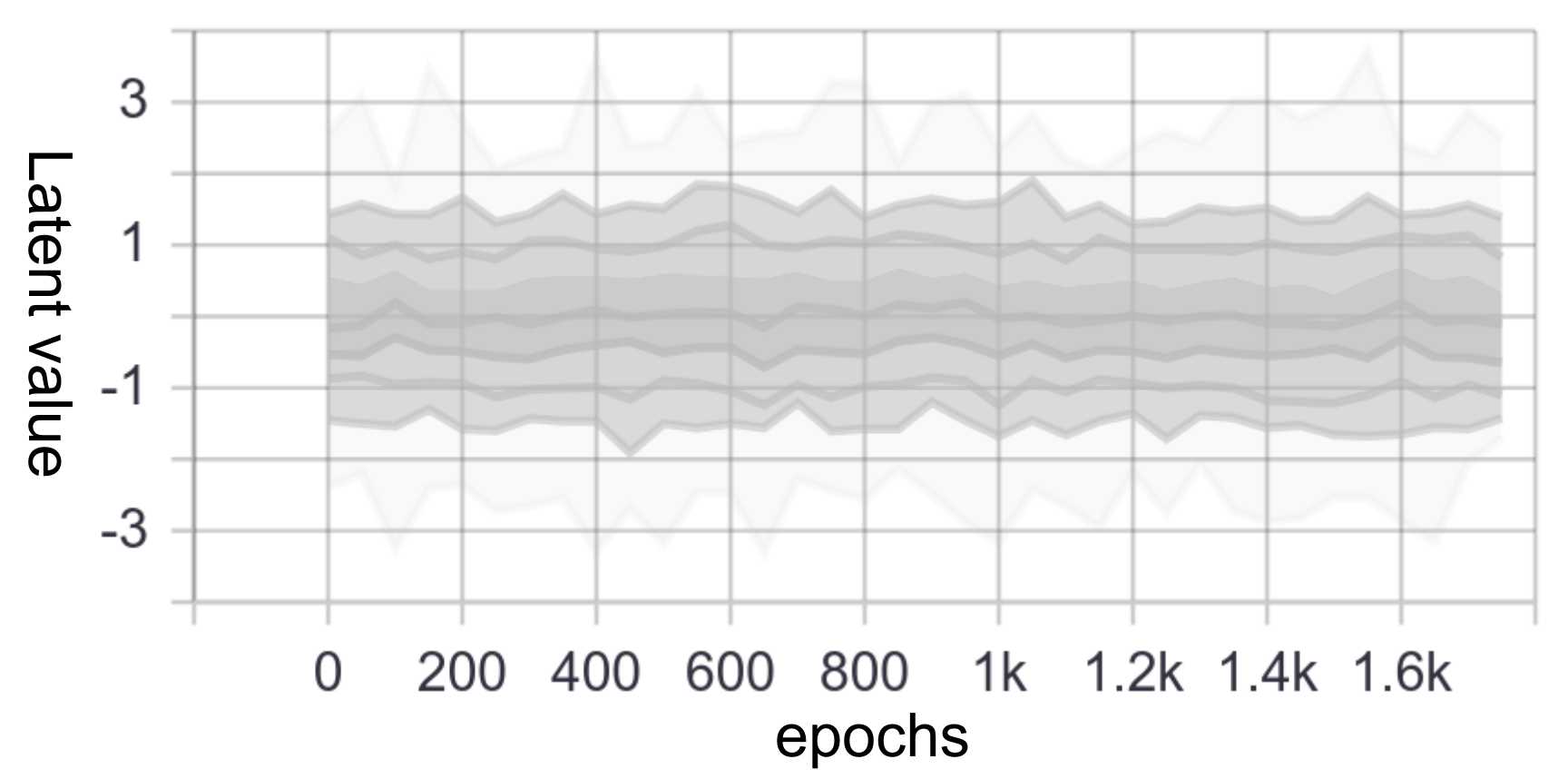}}
    \hspace{5pt}
    \subfloat[The proposed LDL approach]{\includegraphics[width=0.45\linewidth]{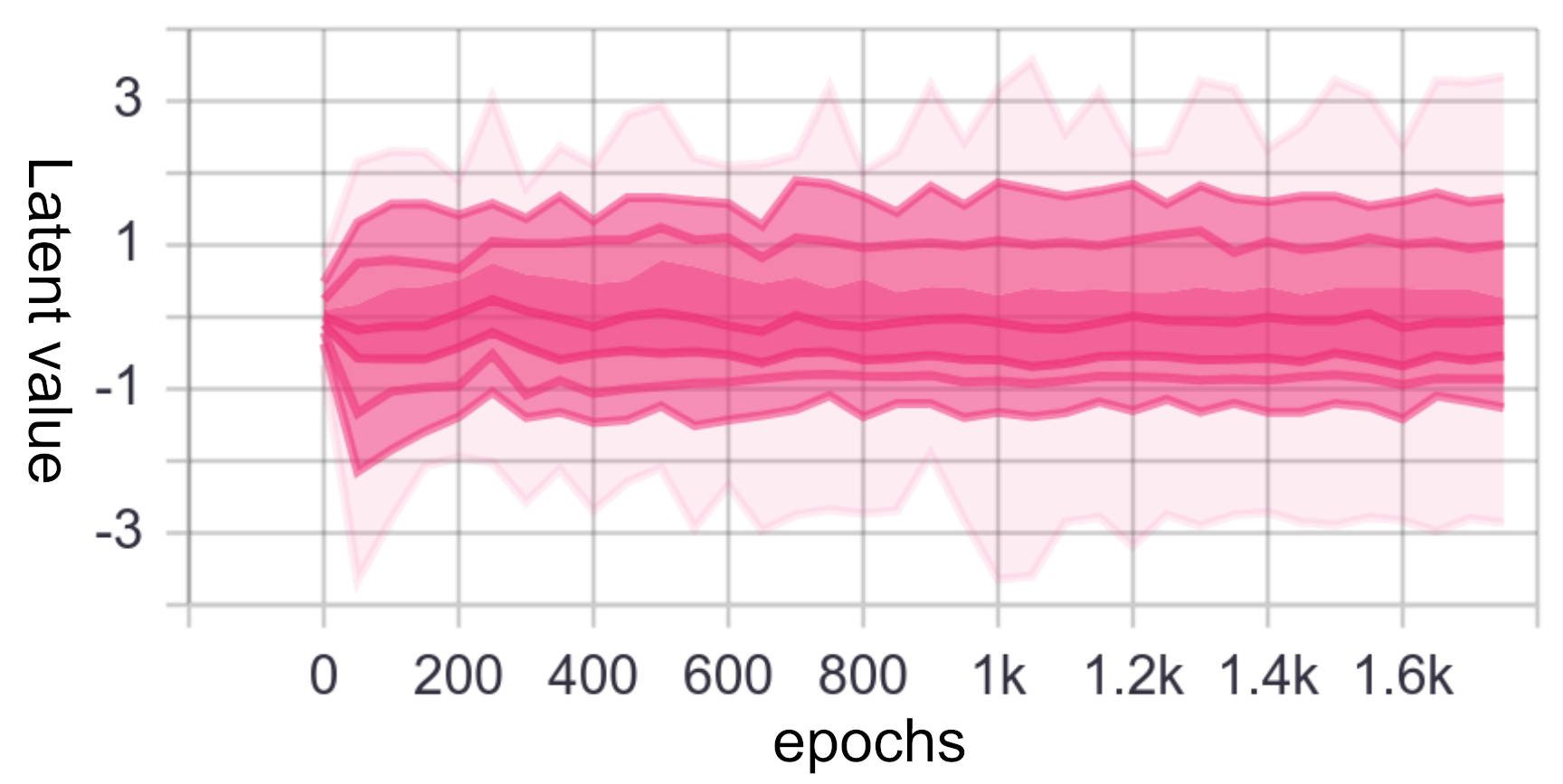}}
    \\
    \subfloat[The proposed AUC approach]{\includegraphics[width=0.45\linewidth]{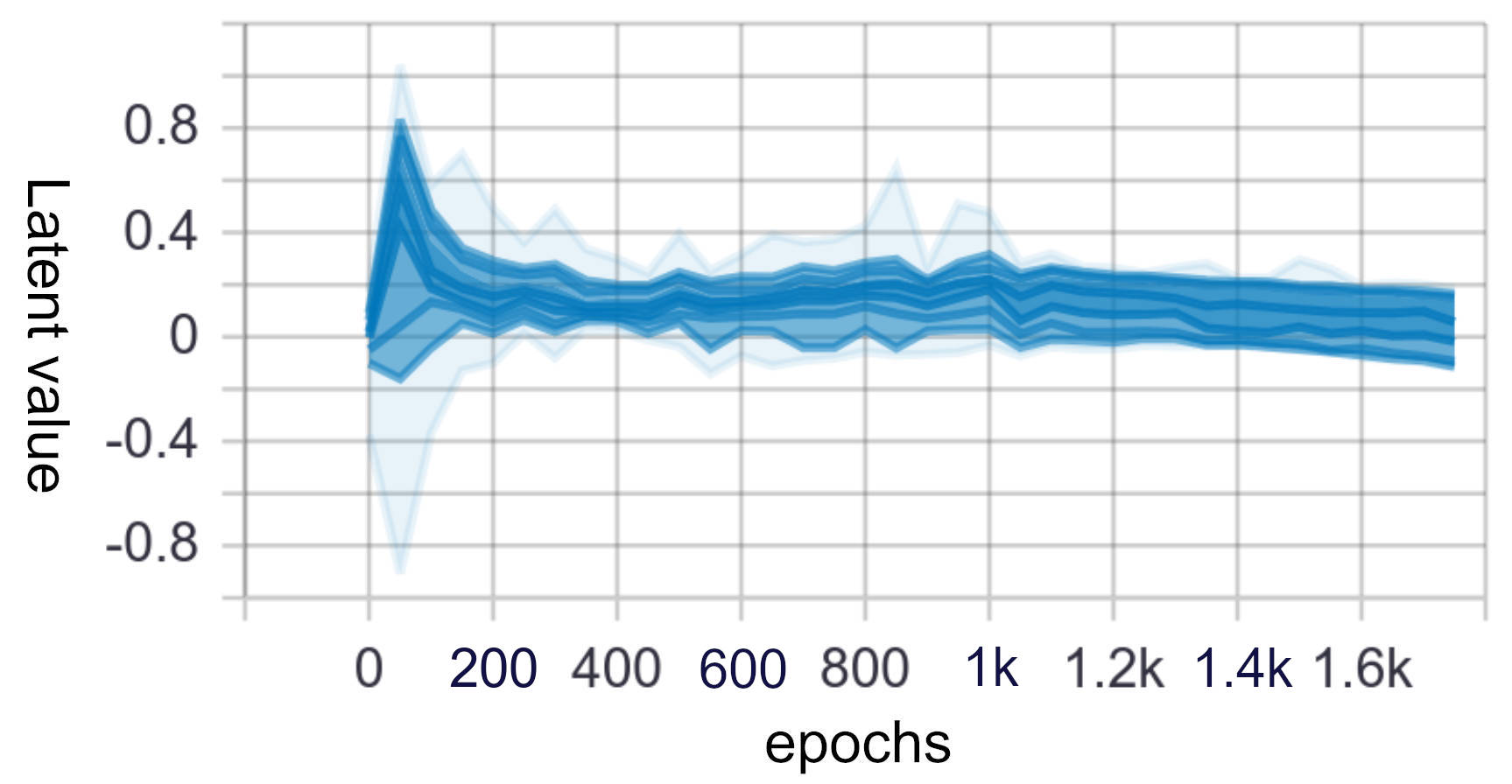}}
    \hspace{5pt}
    \subfloat[The AAE approach]{\includegraphics[width=0.45\linewidth]{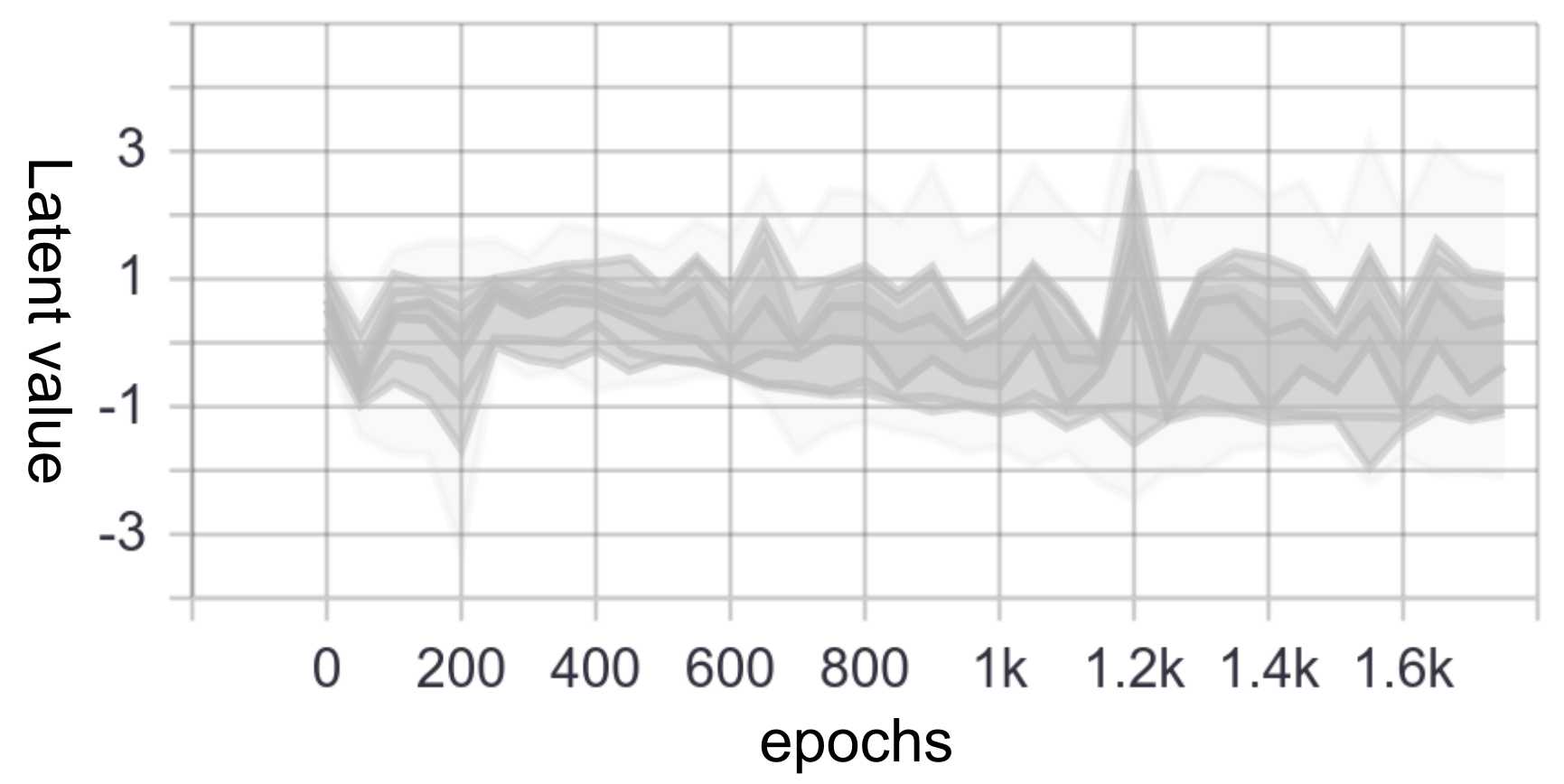}}
    \\
    \subfloat[The VAE approach]{\includegraphics[width=0.45\linewidth]{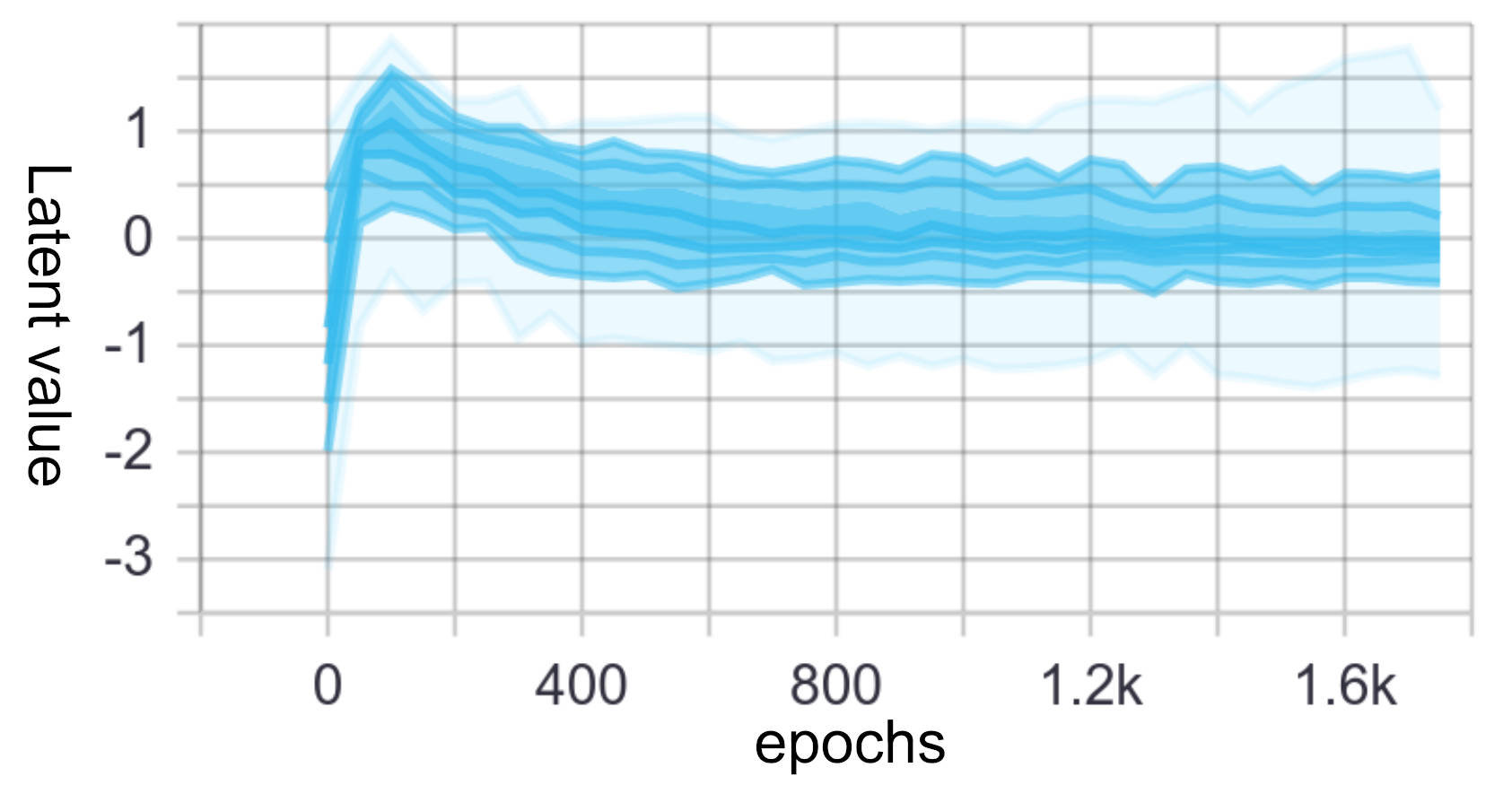}}
    \hspace{5pt}
    \subfloat[The baseline]{\includegraphics[width=0.45\linewidth]{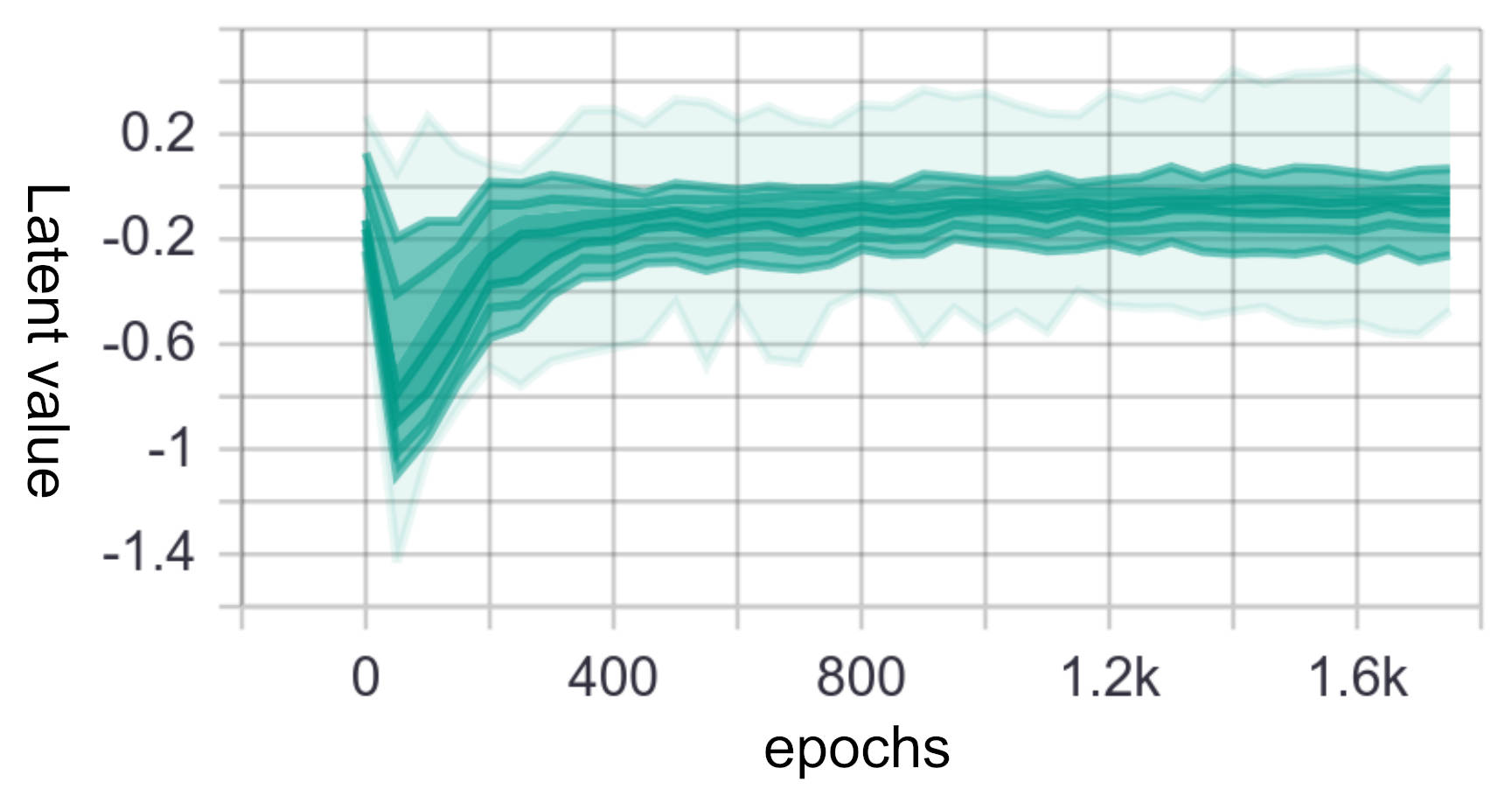}}
    \caption{Visualisation of the (a) objective distribution, compared to the latent distribution developed by the SLR that (b-e) applies four different constraining approaches and (f) imposes no constraint as the training epoch grows.}
    \label{fig:latent-distribution-benchmark}
 \end{figure}

Constraining the latent distribution using four different approaches (i.e., the two proposed approaches, AAE and VAE approaches), the baseline (i.e., without imposing any constraint method), and the objective distribution were compared in this benchmark as shown in Fig. \ref{fig:latent-distribution-benchmark}.

Other than the two proposed approaches, AAE constrains the latent distribution by utilising an adversarial training network. The network employs the encoder in the SLR model as the generator. The discriminator consists of two hidden layers with 100 neurons for each layer and a single neuron for both the input and output layers. The adversarial training process is represented by (\ref{equ:adversarial-training}).

Instead of imposing additional constraints on the latent distribution as used by the previous three approaches, VAE incorporates the posterior distribution of the latent space into the autoencoder network architecture. Since a normal latent distribution is demanded, the encoder outputs two neurons (i.e., a Mean and a Standard Derivation) representing a normal distribution and then generates the single latent variable by sampling the posterior normal distribution.

The last approach imposes no constraint on the latent distribution and is seen as the baseline for the comparisons. 

The objective latent distribution is the standard normal distribution ${\cal N}\left( {0,1} \right)$ as visualised in Fig. \ref{fig:latent-distribution-benchmark}(a), except for the AUC approach.

The results demonstrated that applying the proposed latent learning approach converged rapidly and led to an accurate latent distribution compared to the objective distribution. The proposed supervised learning approach produced a constrictive yet sharp latent distribution. The latent distribution developed by AAE was unstable compared to the rests. Both the distributions developed by VAE and baseline converged slowly during model training with the baseline distribution also being constrictive.

Overall, the proposed latent distribution learning (LDL) approach performed the best. Thus, this approach was selected to constrain the latent distribution in the rest experiments.

\subsection{DoRF curve-fitting benchmark}

\begin{table}[htbp]
    \centering
    \begin{tabular}{c|c|c|c|c}
        \toprule
        \diagbox [width=5em, trim=l] {Method}{Dim} & One  & Two  & Three  & Four  \\
        \midrule
        Our SLR  & \textbf{5.57E-4} & N. A. & N. A. & N. A.     \\
        \hline
        Gamma  & 7.34E-3 & N. A. & N. A. & N. A.     \\
        \hline
        Polynomial  & 8.79E-3 & 2.36E-2 & 2.50E-2 & 2.90E-2    \\
        \hline
        GGCM  & 6.65E-3 & 7.55E-3 & 7.06E-3 & 6.76E-3  \\
        \hline
        EMoR  & 3.60E-2 & 1.24E-2 & 5.71E-3 & 3.21E-3 \\
        \bottomrule
    \end{tabular}
    \caption{DoRF curves fitting performance evaluation of various approximation models with different number of parameters in terms of averaged RMSE.}
    \label{tab:CRF-fitting}
 \end{table}

We firstly compared the performance of the proposed SLR with four other popular models, i.e., gamma, polynomial, GGCM, and EMoR in a DoRF curve-fitting benchmark. In this experiment, every CRF curve in the DoRF was represented by each model with a specific number of parameters using the optimal parameters calculated. Four number of parameters (i.e., 1, 2, 3, and 4) were tested for each of the polynomial, GGCM, and EMoR. While gamma and our method were tested with only one parameter since our method works with a single latent variable. The benchmark results are demonstrated in Table \ref{tab:CRF-fitting}.

The results indicate that our model with only a single parameter achieved greater than a ten-folds better performance (Mean RMSE: 5.61E-4) over most of the other tested methods in the DoRF curve-fitting benchmark. This is not surprising as our model learned the nonlinear CRF features from the real-world CRFs.

\subsection{Camera radiometric calibration}

The performance and applicability of the proposed SLR model is further validated by a camera radiometric calibration application \cite{Grana2004}. This computer vision task estimates inverse-CRF from real camera images.

\begin{figure}[htbp]
    \centering
    \subfloat[Our SLR]{\includegraphics[width=0.45\linewidth]{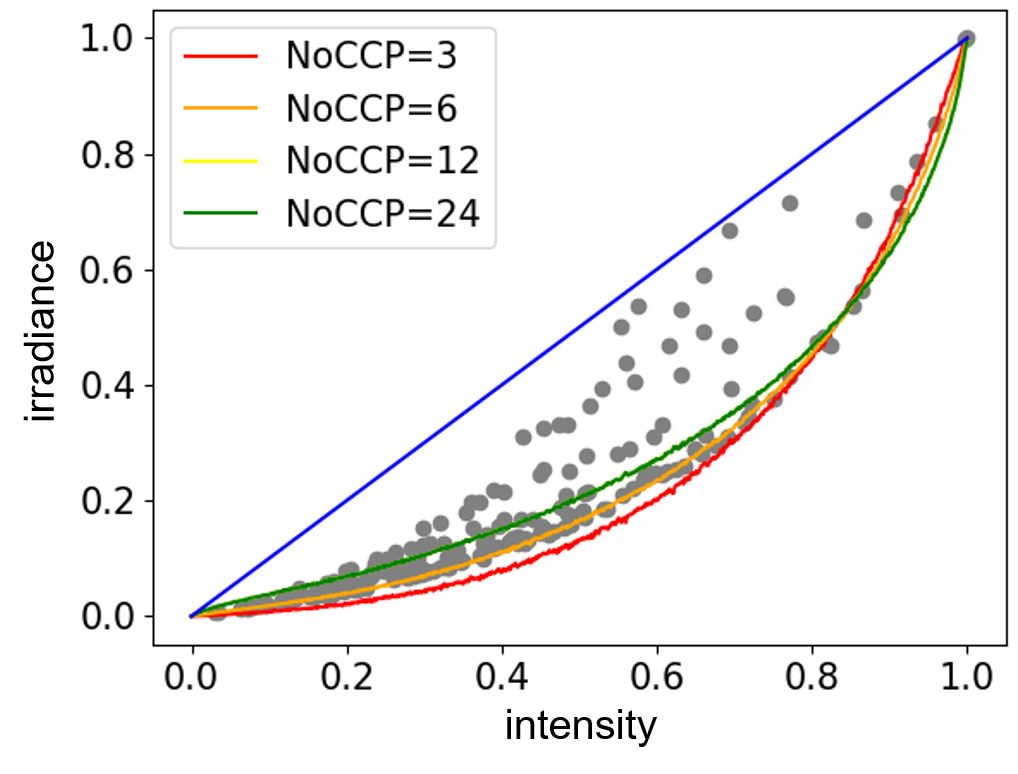}}
    \subfloat[Polynomial]{\includegraphics[width=0.45\linewidth]{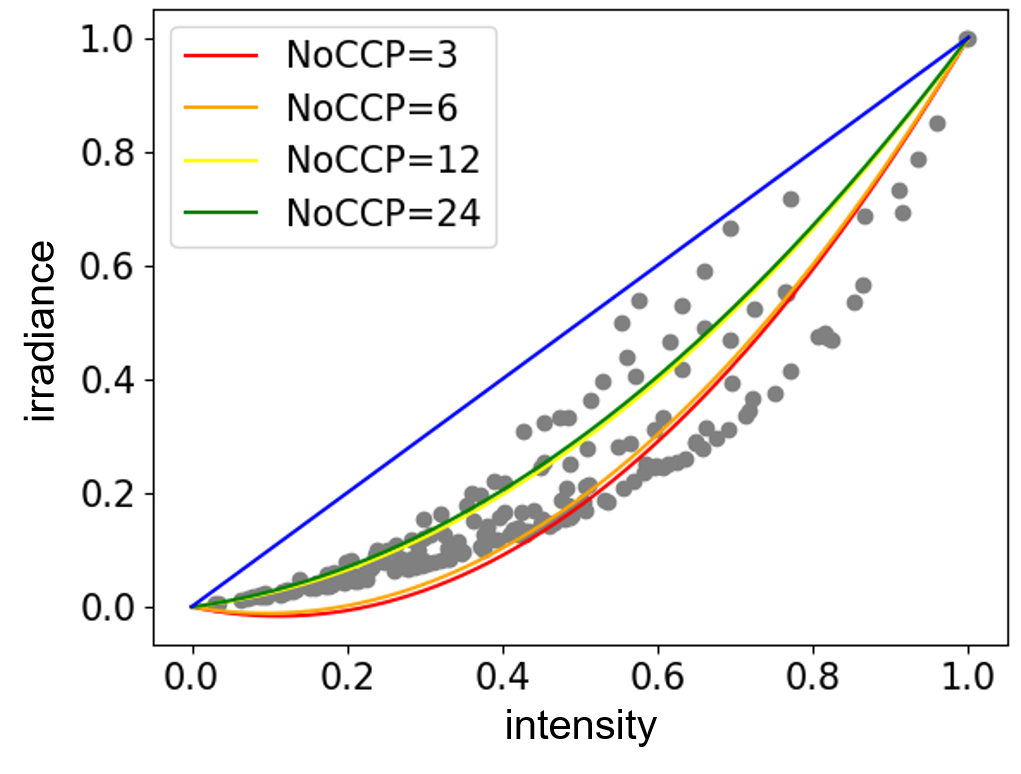}}
    \\
    \subfloat[GGCM]{\includegraphics[width=0.45\linewidth]{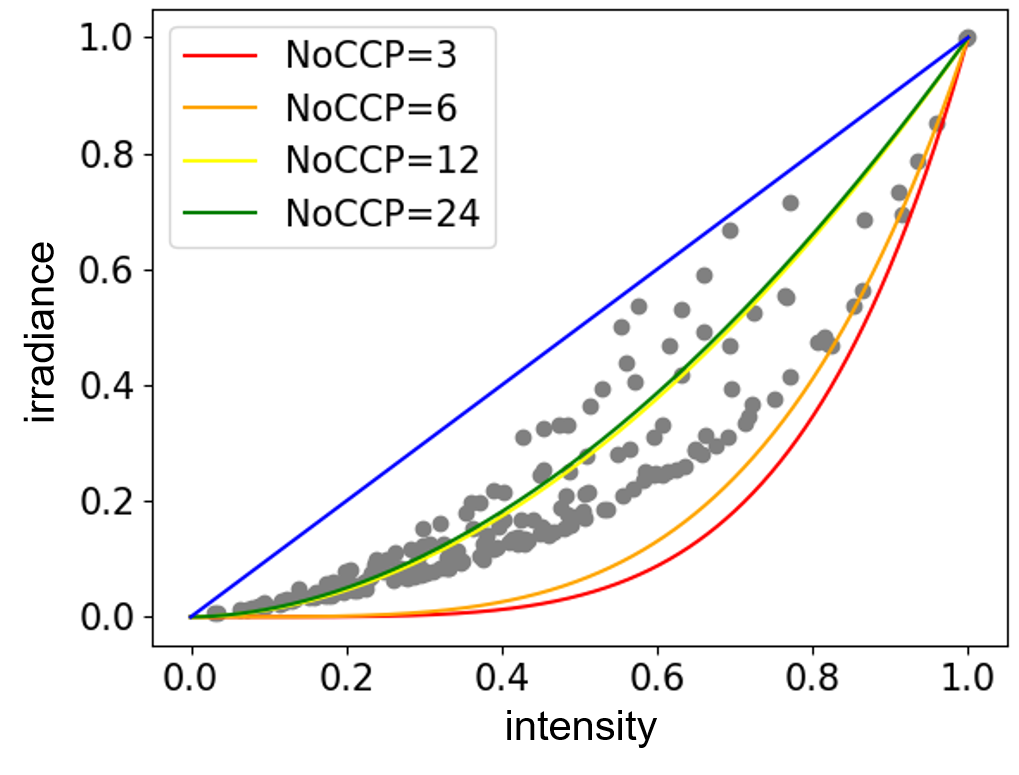}}
    \subfloat[EMoR]{\includegraphics[width=0.45\linewidth]{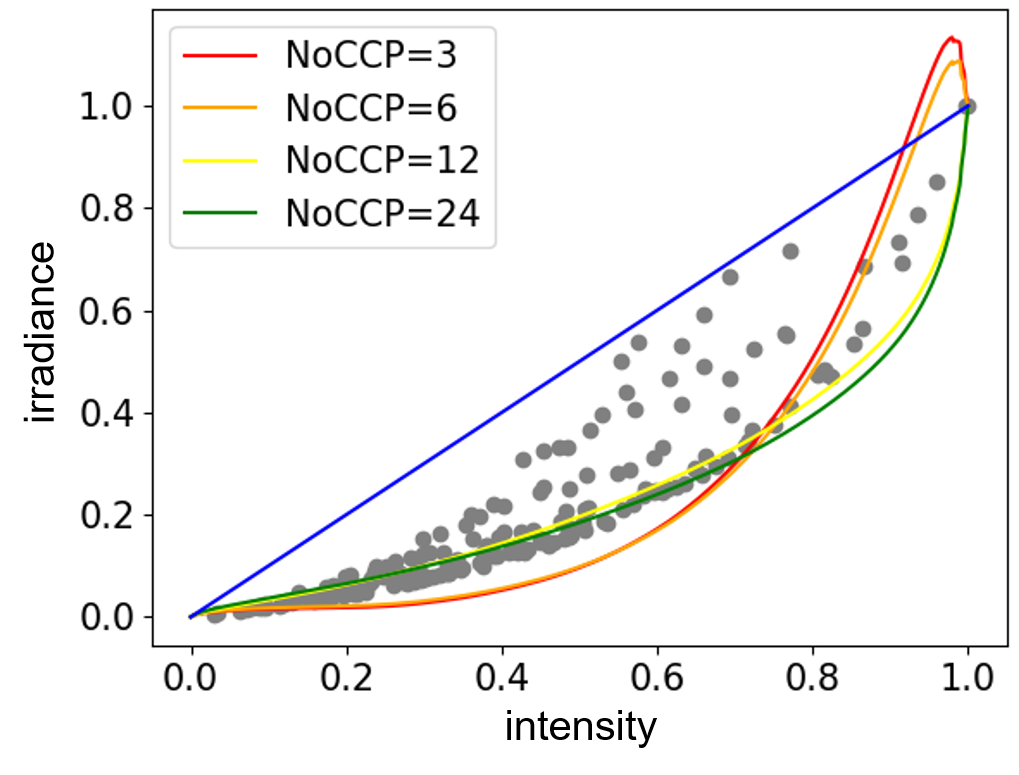}}
    \caption{Radiometric calibration results of a specific camera model (CanonPowerShotG9). The grey dots are the ground truth values. The red, orange, yellow, and red curves are the inverse CRFs calibrated using 3, 6, 12, and 24 NoCCPs on a Macbeth colour chart, respectively. And the blue diagonal line is for reference to. Our model (a) produced more accurate and stable inverse CRFs than the other three tested methods (b, c, d).}
    \label{fig:calibration}
 \end{figure}
 
 \renewcommand{\arraystretch}{1.5}
 \begin{table}[htbp]
    \footnotesize
    \centering
    \begin{tabular}{c|c}
       \toprule
       \scriptsize{Method}  &  \scriptsize{Stability (Total Variance) $\downarrow$}  \\
       \hline
       Our SLR  & \textbf{0.66}  \\
       \hline
       Polynomial  & 1.63   \\
       \hline
       GGCM  & 8.11  \\
       \hline
       EMoR  & 3.87   \\
       \bottomrule
    \end{tabular}
    \caption{Stability evaluation and comparison of the four commonly used CRF models, i.e., our SLR, polynomial, GGCM, and EMoR, in terms of the total variance between CRFs estimated using 3, 6, 12, and 24 NoCCPs on a Macbeth colour chart.}
    \label{tab:crf-stability}
\end{table}
\renewcommand{\arraystretch}{1}

The true irradiance-intensity mapping values of a specific camera model, i.e., Canon PowerShot G9, and inverse-CRFs produced by four different methods with applying 3, 6, 12, and 24 number of corresponding colour patches (NoCCPs) during calibration are visualised in Fig. \ref{fig:calibration}. Our model fitted more accurately to the true values (see Table \ref{tab:crf-stability} for detail). Ours also performed more stable when using varied NoCCPs for the calibration (the Total Variance of the four curves in each plot of Fig. \ref{fig:calibration} are: our SLR 0.66; polynomial 1.63; GGCM 8.11; EMoR 3.87; the smaller the better).

\renewcommand{\arraystretch}{2}
\setlength\tabcolsep{1pt}
\begin{table}[ht]
    \footnotesize
    \centering
    \begin{tabular}{c|c|c|c|c|c|c|c}
        \toprule
        \multicolumn{2}{c|}{}  & \footnotesize {Mean}  & \footnotesize {Median}  & \footnotesize {S.D.}   & \footnotesize {Max. }   & \footnotesize {95 Pct.} & \footnotesize {Time} \\
        \midrule
        \multirow{4}{*}{\makecell[c]{Our SLR\\ (d=1)}} & Our LDL & \textbf{0.062} & \textbf{0.056}   & \textbf{0.056} & \textbf{0.106}  & \textbf{0.101}   & 57.4   \\
        & Our AUC  & 0.105 & 0.092   & 0.064 & 0.252  & 0.213   & \textbf{43.1}   \\
        & AAE  & \textbf{0.064} & \textbf{0.057}   & \textbf{0.027} & 0.131  & \textbf{0.107}   & 56.4   \\
        & VAE  & \textbf{0.063} & \textbf{0.054}   & \textbf{0.025} & \textbf{0.108}  & \textbf{0.107}   & 57.0   \\
        & baseline  & 0.075 & 0.067   & 0.039 & 0.193  & 0.138   & 58.5   \\
        \hline
        \multirow{4}{*}{\makecell[c]{Our SLR\\ (d=3)}} & Our LDL & 0.097 & 0.097   & 0.050 & 0.184  & 0.181   & 61.1   \\
        & AAE  & \textbf{0.063} & \textbf{0.051}   & 0.038 & 0.178  & 0.122   & 60.6   \\
        & VAE  & 0.091 & 0.091   & 0.038 & 0.145  & 0.139   & 62.1   \\
        & baseline  & 0.120 & 0.120   & 0.051 & 0.251  & 0.195   & 59.72   \\
        \hline
        \multicolumn{2}{c|}{Gamma (d=1)}  & 0.092 & 0.080 & 0.040 & 0.164 & 0.162 & 112.6   \\
        \hline
        \multicolumn{2}{c|}{Polynomial (d=3)}  & 0.074 & 0.070 & \textbf{0.027} & 0.156 & 0.118 & 129.6   \\
        \hline
        \multicolumn{2}{c|}{GGCM (d=3)}  & 0.124 & 0.125 & 0.055 & 0.222 & 0.210 & 143.3  \\
        \hline
        \multicolumn{2}{c|}{EMoR (d=3)}  & 0.154 & 0.105 & 0.121 & 0.474 & 0.408 & 138.1 \\
        \bottomrule
    \end{tabular}
    \caption{Camera radiometric calibration results produced by five different methods (our SLR, gamma, polynomial, GGCM, and EMoR) using eight calibration images and three colour patches in each image are listed. Our SLR was evaluated with four latent distribution constraining approaches and the baseline. Six metrics are used to evaluate the performance of each method. The first five are statistical metrics (mean, median, standard derivation, maximum, and 95 percentile) of RMSE that reflect model accuracy. Among these five metrics, the mean can be seen as the overall performance metric for accuracy. The time metric was evaluated as the total time needed in seconds for calibrating all the 14 camera models.  }
    \label{tab:calibration-results}
\end{table}
\renewcommand{\arraystretch}{1}

The radiometric calibration performances of the five methods (i.e., our SLR, gamma, third degree polynomial, third degree GGCM, and EMoR with three parameters) were further evaluated by 14 camera models. Their performances in terms of six metrics are demonstrated in Table \ref{tab:calibration-results}. The first five metrics are statistical metrics of the RMSEs calculated from the inverse-CRFs of the 14 camera models. The RMSE of each camera model was calculated by comparing the true values and the calibrated inverse-CRF. These five metrics evaluate accuracy of the inverse-CRFs calibrated by each method. Our SLR with a single latent variable and applying the LDL (Mean RMSE 0.062) clearly outperformed the others compared to some of the other methods with using even three parameters. The CRF calibration accuracy improvement contributed by the LDL on our SLR can be quantified by comparing with the baseline. The sixth metric evaluates the total time needed for calibrating all the 14 camera models (i.e., finding the optimal model parameters). It is a metric that reflects model efficiency and is important to be considered for deploying on mobile platforms. Our SLR with LDL (57.4s) completed all the calibrations almost twice faster than the gamma (112.6s) that also works with a single parameter and the others that work with more parameters. This is partially contributed by the simple yet efficient autoencoder architecture found by NAS. Our SLR with AUC (43.1s) achieved faster calibration yet sacrificed the calibration accuracy (Mean RMSE 0.105).

\section{Conclusion}\label{sec13}

In this paper, a CRF model that represents camera responses with only a single latent variable has been described. The model used unsupervised learning on real-world CRFs by autoencoder. A simple yet efficient autoencoder architecture was found by applying a naïve NAS algorithm. A latent distribution learning approach was introduced to effectively constrain the latent variable to a normal distribution for improving the accuracy of the CRF calibration process. We demonstrated a superior performance of the proposed model in terms of both the CRF modelling accuracy (i.e., a ten-folds better CRF modelling accuracy in the curve-fitting cross validation benchmark) and calibration efficiency (i.e., around twice as fast as the best current models for CRF calibration in a double-cross validation benchmark). 

\section*{Declarations}

\bmhead{Funding}
This project was supported in part by the European Union’s Horizon 2020 research and innovation program under the Marie-Sklodowska-Curie grant agreement No 720325, FoodSmartphone, and in part by the Key Laboratory of Intelligent Preventive Medicine of Zhejiang Province 2020E10004.

\bmhead{Availability of code}
Code is available at https://github.com/zyfccc/Representing-camera-response-function.



\input sn-sample-bib.tex%

\end{document}